\documentclass[11pt]{article}

\usepackage{graphicx}
\usepackage{float}
\usepackage{color}
\usepackage{amsmath,amsthm,amssymb,multirow,paralist} 
\usepackage{subcaption}
\usepackage[english]{babel}

\usepackage[margin=0.8in]{geometry}
\usepackage{hyperref}

\begin{document}

\begin{center}
{\Large \textbf{Emotion detection from social media posts}}

\vspace{10pt}
\text{Md Mahbubur Rahman, Shaila Sharmin}

mdrahman@iastate.edu, ssharmin@iastate.edu 

Iowa State University

\end{center}




\begin{abstract}

Over the last few years, social media has evolved into a medium for expressing personal views, emotions, and even business and political proposals, recommendations, and advertisements. We address the topic of identifying emotions from text data obtained from social media posts like Twitter in this research. We have deployed different traditional machine learning techniques such as Support Vector Machines (SVM), Naive Bayes, Decision Trees, and Random Forest, as well as deep neural network models such as LSTM, CNN, GRU, BiLSTM, BiGRU to classify these tweets into four emotion categories (Fear, Anger, Joy, and Sadness). Furthermore, we have constructed a BiLSTM and BiGRU ensemble model. The evaluation result shows that the deep neural network models(BiGRU, to be specific) produce the most promising results compared to traditional machine learning models, with an 87.53 $\%$  accuracy rate. The ensemble model performs even better (87.66 $\%$), albeit the difference is not significant. This result will aid in the development of a decision-making tool that visualizes emotional fluctuations.

\end{abstract}

\section{Introduction}

Emotions are a cognitive/mental state influenced by  neurophysiological changes in humans. Our behavioral responses vary depending on our mental and psychological states; it might be pleasure or unhappiness or other factors that significantly impact our daily activities. Fear, anger, joy, and sadness are all examples of emotional states that affect our mental health and decision-making ability. People nowadays use the internet and social media platforms like Facebook and Twitter to express their thoughts, feelings, and opinions. On social media, millions of people share, discuss, upload, and comment on every event, action, or news that occurs across the world. Therefore, emotion analysis is critical for understanding an individual's emotional state of mind because it shows a person's true voice and perspective,  detects any potential future issues and offers solutions for assisting an individual with anxiety. It can also be used to determine a person's mental or emotional state by evaluating his or her behaviors over time and recognizing depression risks.

This emotion classification challenge of our project aims to detect and identify different sorts of feelings through the form of texts collected from Twitter. 
SVM, Naive Bayes, Decision Tree, Random Forest, and Logistic Regression are some examples of traditional machine learning algorithms that have a good reputation for evaluating and categorizing a large amount of text data in a short amount of time \cite{nabeel2021classifying, ahuja2022sentiment}. We started with these models at first, but as time went on, we discovered that deep neural networks (LSTM, CNN, GRU, and others) function even better at capturing the sentiment associated with the pos since we want to create a model that can accurately categorize emotions from the text for unknown data.

We used a labeled dataset of around 7000 samples with four different emotion classes (fear, anger, joy, and sadness) collected from kaggle.com. However, we have pre-processed the dataset before training the above-mentioned machine learning methods. After that, we evaluated each of the trained models with test data. The datasets utilized for classification tasks are quite large, allowing for more comprehensive coverage of the vocabulary that may be found on Twitter or other social media sites.

In this work, we aim to develop the optimal machine learning model to predict emotions from unknown text data with acceptable accuracy by finding the best hyper-parameters that may operate well in our dataset. Moreover, we aim to develop decision-making to depict emotional variations in human beings and suggest helpful content for their social media accounts.

\section{Related Work}
Throughout the year, numerous research have been conducted to classify  positive and negative attitudes in reviews, ratings, comments and suggestions for movie or products ,and other sources of online expression. Emotions have also been explored; however, it still is a rising research topic. Bharat Gaind et al. \cite{gaind2019emotion} proposed a method for classifying and quantifying tweets based on Paul Ekman's six standard emotions using tweets from Twitter \cite{ekman1992argument}. Their system can label and score any piece of text data, especially tweets, as well as build an efficient training dataset automatically for ML classifiers. 

Another work on this topic was done by S. X. Mashal, and K. Asnani \cite{mashal2017emotion}. They use machine learning techniques (Nave Bayes and SVM) to perform emotion analysis and classification. They also used Twitter data by using emotion hashtags for five different emotion categories: happy, sad, angry, fear, and surprise. This study also shows how changing the size of the training data has an impact. The impact of sarcastic expression on social appearance and perception is another challenging factor. A framework for investigating the overlap of emotions in a text as a defining feature of sarcastic phrases was also conducted by Fernando H Calderon et al. \cite{calderon2019emotion}.

Another study was conducted for emergency management to monitor what individuals post online and alter their information methods to reflect the expectations and needs of their target audience\cite{brynielsson2014emotion}.

All of the studies described above use different machine learning algorithms as well as focus on creating a new dataset and labeling them; however, none of them evaluated the hyper-parameters for each of the machine learning techniques as explicitly as we have considered.

\section{Methodology}

Initially, we collected the data from Kaggle for this project. After pre-processing the data, we trained some machine learning and deep learning classifiers using the train dataset. Finally, we have evaluated each of these models using the test dataset. 

\subsection{Data Collection}

We downloaded the data directly from Kaggle.com\footnote{https://www.kaggle.com/datasets/anjaneyatripathi/emotion-classification-nlp}. The dataset consists of four different classes: joy, anger, fear and sadness. We consider the tweets that are written English. The dataset contains a total of 7102 data samples . Table \ref{tab:dataset} Shows the distribution of data for different emotion categories.
\begin{table}[!ht]
\centering
\caption{Data distribution among different classes}
\begin{tabular}{|c|c|c|c|c|}\hline
\textbf{Classes}  & anger & joy & sadness & fear\\ \hline
\textbf{Size}  & 1701 & 1616 & 1533 & 2252\\ \hline
\end{tabular}
\label{tab:dataset}
\end{table}

\subsection{Data Pre-Processing}

To begin with pre-processing of data, we removed any punctuation, HTML tags, URL links, and stop words. Each abbreviated phrase was converted to its original form. For exam 'omg' or 'ASAP' is converted to oh my god and as soon as possible. We then converted texts into the lowercase letter. We have also used lemmatization and stemming to reduce our vocabulary and remove the redundancy of the same types of words to find the root words.  We keep the emoticons associated with the post as it is since it also involves in revealing expression of current state of mind.

We used 80\% of the data for training and validation purpose and the rest of 20\% of the dataset was used for testing purpose.  From the training and validation dataset, 10\% of the data was used for validation purpose.

\subsection{Feature Selection and Extraction}

We construct a vocabulary by collecting all of the words that appear throughout the entire pre-processed dataset. The feature set is made up of this vocabulary. We tested two alternative strategies for feature extraction: word embedding and bag-of-words.  We used the pre-trained word embedding with embedding size 300 which was trained on Google News Data\footnote{https://code.google.com/archive/p/word2vec/}. During training, we fine-tuned the word embedding using our training dataset.

Although, surprisingly bag-of-words shows better performance compared to word embedding for this dataset in some traditional machine learning classifiers, it performs poorly in deep neural networks whereas word embedding shows prominent result. On the other hand, we can not use bag of words as a feature extractor in recurrent neural network(RNN) models like LSTM, BiLSTM, GRU, BiGRU because the order of words matter in these models.

Therefore, we found two things: (1). Some of the traditional machine learning models perform better with bag-of-words but the performance gap among the traditional machine learning algorithm is too high. For example, logistic regression achieved accuracy 81.34\% while K-NN shows accuracy 41.50\%. (2). We can't use bag-of-words with the RNN models. Also, the performance gap among the deep learning model is very low and it indicates that it is better to use word-embedding instead of bag-of-words.

So, considering the above two points,  we finally selected word embedding  technique as our feature extraction technique. 

\begin{table}[!ht]
\centering
\caption{Performance analysis of word embedding and bag-of-words. }
\begin{tabular}{|c|c|c|c|}\hline

Models	& Accuracy(Word Embedding) & Accuracy(Bag of Words)\\ \hline
Logistic Regression	& 0.6952 & 0.8134 \\ \hline
Naive-Bayes & 	0.2449 & 0.582 \\ \hline
Support Vector Classifier &	0.6892 & 0.7692 \\ \hline
K-Neighbors Classifier &	0.4212 & 0.4150 \\ \hline
Decision Tree Classifier &	0.4804 & 0.7645\\ \hline
Random Forest Classifier &	0.5829 & 0.8154\\ \hline
CNN &	0.8568 & N/A \\ \hline
LSTM &	0.8596 & N/A \\ \hline
BiLSTM &	0.8601 & N/A \\ \hline
GRU &	0.8624 & N/A \\ \hline
BiGRU &	0.8712 & N/A \\ \hline
\end{tabular}
\label{tab:fe}
\end{table}

Table \ref{tab:fe} shows the performance between bag-of-words and word embedding  with different methods. The performance of each of the models is based on default hyper-parameters. We didn't do any hyper-paramter tuning during this section. The default hyper-parameters for traditional machine learning are based on  sklearn library and we consider batch size 32 and learning rate 0.001 as defaut hyper-parameters for the deep learning models. We ran each of the models three times and calculated the accuracy by taking the average of three accuracy scores. 

\subsection{Traditional Machine Learning Classifiers}

Some traditional machine learning algorithms such as Logistic Regression, Support Vector Machine, Naive Bayes, K-NN, Decision Tree, and Random Forest classifiers are used in this project to solve the multi-class classification problems in our data \cite{chougule2022machine}. These models are trained and evaluated using the Sklearn library \footnote{hrefhttps://scikit-learn.org/stable/https://scikit-learn.org/stable/}.

\subsection{Deep Learning Classification Methods}

We trained LSTM, BiLSTM, GRU, BiGRU, and Ensemble of (BiGRU+BiLSTM) deep learning models. These models are trained and evaluated using three different types of architecture.

\begin{figure}[H]
      \centering
        \includegraphics[width=19cm, height=8cm]{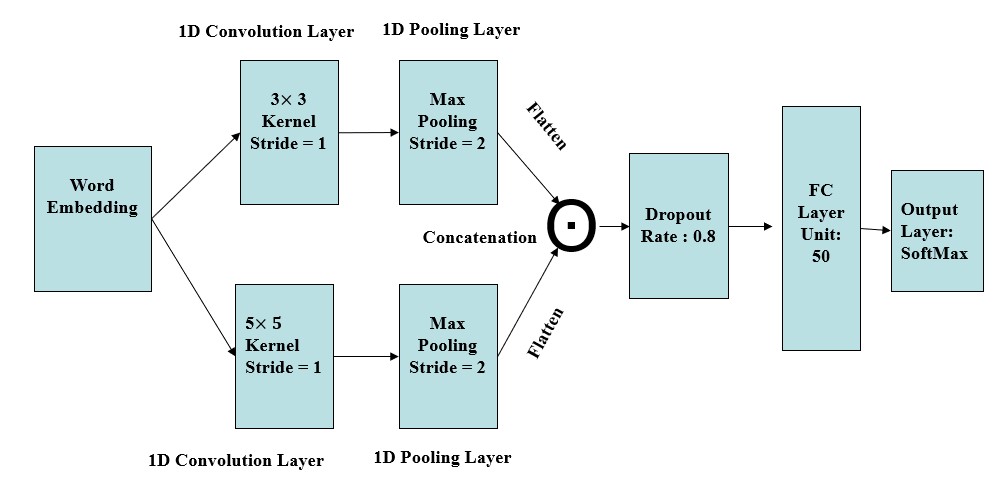}
      \caption{Architecture for Convolutional Neural Network (CNN)}
      \label{fig:cnn_model}
  \end{figure}

For the CNN model, we adopted the architecture from Yoon Kim's paper "Convolutional Neural Networks for Sentence Classification". \cite{kim-2014-convolutional} \footnote{\href{https://github.com/yoonkim/CNN\_sentence}{https://github.com/yoonkim/CNN\_sentence}} The architecture of the model is shown in figure \ref{fig:cnn_model}. 
 
The tokenized data is used as the input to the word embedding layer. The output of the word embedding layer is then sent to two convolution layer parallelly. One convolution layer uses kernel of size $3 \times 3$ and another layer uses kernel of size $5 \times 5$. Both of them use stride size 1. After each of the convolution layer, there is a max pooling layer of stride size 2. The outputs of the two max pooling layer is then concatenated followed by a dropout layer with a dropout rate of 0.8. Furthermore, there is a fully connected layer with 50 units. The output of the FC layer is finally forwarded to the output/SoftMax layer.

\begin{figure}[H]
      \centering
        \includegraphics[width=18cm]{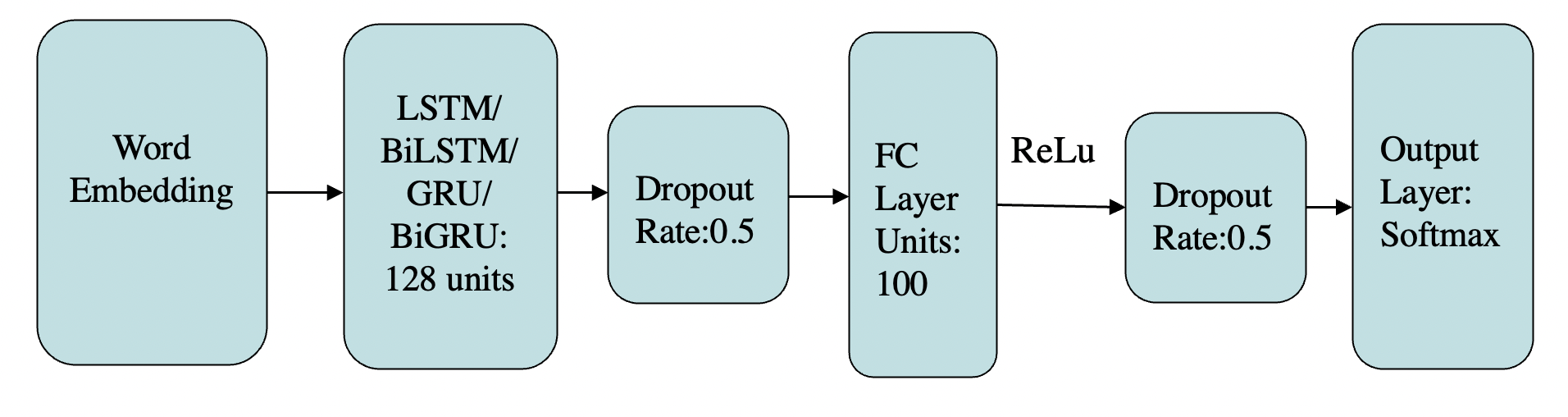}
      \caption{Architecture for LSTM/GRU/BiLSTM/BiGRU}
      \label{fig:rnn_models}
  \end{figure}

Figure \ref{fig:rnn_models} shows the architecture for LSTM, BiLSTM, GRU, and BiGRU and they share the same architecture.

  At first we begin with the word embedding layer with embedding size of 300. Then, there is an LSTM/BiLSTM/GRU/BiGRU layer of 128 units followed by a dropout layer with a dropout rate of 0.5. After that, a fully connected layer of 100 neurons is added followed by the activation function ReLU. Again, we have a dropout layer with a dropout probability of 0.5. Finally, we have the softmax layer as our output layer.

  For the ensemble model, we trained BiLSTM and BiGRU using the above architecture and then combined the result by the majority rule.

\subsection{Training and Hyper-Parameter Tuning}

\subsubsection{Training}

Traditional machine learning methods are trained using the Sklearn library, while deep learning models are trained using the Keras library. Google Colab \cite{carneiro2018performance} is used for all of the training. We used colab's GPU during training the deep learning models.

Before training traditional machine learning models, we first find the embedding vector of each word of the text using the fine-tuned word embeddings and then stack the vectors of all the tokens/words to produce a matrix. The matrix is then flattened and used as the input for these models.

The word embedding layer in deep learning models only takes the token id of the words/tokens as input and finds the word embedding. The weights are fine-tuned during training, and the embedding layer is initialized using the pre-trained word embedding.

 \subsubsection{Hyper-Parameter Tuning}
 To acquire the optimal hyper-parameter settings for each of the models, we performed  hyper-parameter tuning using the training data set.

\begin{table}[!ht]
\centering
\caption{Hyper-parameters that were used in tuning and the best hyper-parameters that were chosen for traditional machine learning models}
\begin{tabular}{|c|p{8cm}|p{4cm}|}\hline
\textbf{Models}	& \textbf{Hyper-parameters} & \textbf{Best Hyper-parameters} \\ \hline
    Logistic Regression	& C = \{ 1000, 10, 1, 0.1, 0.01, 0.001\}

    penalty = \{l1, l2\}

    solver =  \{liblinear,lbfgs, saga, newton-cg, sag\} & 
    C = 10

    solver = liblinear

    penalty = l2
   \\ \hline
    Support Vector Classifier &	
    C = \{ 10, 1, 0.1\} 
    
    gamma= \{1, 0.1, 0.001, 0.01\}
    
    kernel= \{sigmoid, poly, rbf ,  linear\}
    & C= 1
    
    gamma= 0.1
    
    kernel= rbf
    \\ \hline
    K-Neighbors Classifier &	
    leaf\_size = 1 - 50
    
    n\_neighbors = 1 - 30
    
    p=\{1,2\}
    & 
    p= 2
    
    n\_neighbors= 3
    
    leaf\_size= 29
    \\ \hline
    Decision Tree Classifier &	
    max\_depth = \{None, 15, 30\}
    
    min\_samples\_split = \{10, 2, 5, 15\}
    
    min\_samples\_leaf = \{5, 1, 2\} 
    & 
    max\_depth = None
    
    min\_samples\_leaf= 1 
    
    min\_samples\_split = 10
    \\ \hline
    Random Forest Classifier &	
    max\_depth= \{None, 15, 30\}
    
    n\_estimators= \{200, 100, 300\} 
    
    min\_samples\_leaf = \{5, 2, 1\}
    
    min\_samples\_split = \{15, 10, 2, 5\}
    &  max\_depth = None
    
    n\_estimators = 300
    
    min\_samples\_split = 15
    
    min\_samples\_leaf = 2
    \\ \hline

\end{tabular}
\label{tab:hyper-params}
\end{table}

For the traditional machine learning models, we used GridSearchCV from Sklearn \footnote{\href{https://scikit\-learn.org/stable/modules/generated/sklearn.model\_selection.GridSearchCV.html}{https://scikit\-learn.org/stable/modules/generated/sklearn.model\_selection.GridSearchCV.html}} to do the hyper-parameter tuning. We performed 10 folds cross-validation techniques on the training and validation data during the hyper-parameter tuning. Table \ref{tab:hyper-params} lists
the various hyper-parameters that were utilized in tuning and the best hyper-parameter that were chosen
in the end for the traditional machine learning models.

We fine-tuned  learning rate, batch size, and the number of epochs for deep learning models. In order to make a fair comparison, we need to use the same hyper-parameters in all deep learning models. That is why we fine-tuned the hyper-parameters only for the LSTM model, and we found batch size 16 and learning rate 0.001 as the best hyper-parameters. As a result, we used batch size 16 and a learning rate of 0.001 to train all of the models. Because we found no significant performance improvement between 25 and 35 epochs, all models are trained until 35 epochs. Table \ref{tab:hyper-params} lists the various hyper-parameters that were utilized in tuning and the optimal hyper-parameter that were chosen
in the end for the deep learning models.

\begin{table}[!ht]
\centering
\caption{Hyper-parameters that used in tuning and the best hyper-parameters that were selected for deep learning models}
\begin{tabular}{|c|c|c|c|c|}\hline
\textbf{Batch-size}  & 8, 16, 32, 64, 128\\ \hline
\textbf{Learning Rate}  & 0.1, 0.01, 0.001, 0.0001 \\ \hline
\textbf{Best Accuracy(LSTM)}  & \textbf{0.8619} (Batch:16, Learning-rate: 0.001)\\ \hline
\end{tabular}
\label{tab:hyper-parameter_deepLearning}
\end{table}

\section{Experimental Result}

After finding the appropriate hyper-parameters, we evaluated them on the test data. Table \ref{tab:result} shows the evaluation results for all of the models. We observed that the ensemble model of BiLSTM and BiGRU surpasses all other models with an accuracy rate of $87.66\%$. However, other deep learning models such as LSTM, CNN, BiGRU, BiLSTM, and CNN  also achieved accuracy close to the ensemble model in classifying/detecting emotions in texts compared to all of the traditional machine learning models.

\begin{table}[!ht]
\centering
\caption{Accuracy of different models using the best hyper-paratmeters }
\begin{tabular}{|c|c|c|c|}\hline
\textbf{Traditional Models}	& \textbf{Accuracy} & \textbf{Deep Learning Models} & \textbf{Accuracy} \\ \hline
Logistic Regression	& 0.6953 & CNN & 0.8601\\ \hline
Naive-Bayes & 	0.2449 & LSTM & 0.8619 \\ \hline
Support Vector Classifier &	0.6892 & BiLSTM & 0.8630 \\ \hline
K-Neighbors Classifier & 0.4212 & GRU & 0.8645\\ \hline
Decision Tree Classifier &	0.4804 & BiGRU & 0.8753\\ \hline
Random Forest Classifier &	0.5829 & \textbf{Ensemble (BiGRU + BiLSTM)} &\textbf{0.8766}\\ \hline

\end{tabular}
\label{tab:result}
\end{table}

Naive Bayes, on the other hand, performs poorly among traditional machine learning models. In this project, words are being considered as features that could be dependent on each other to express specific emotions in a text. Since Naive Bayes strongly assumes that all the features are independent given the output class, it fails to capture the dependency among the words and hence performs poorly.

However, Nearest neighbor, on the other hand, is heavily reliant on point distances. The dimensionality curse occurs when the number of dimensions is large, and distances become less representative. This results in poor performance. The dimension of each text data point in our project is equivalent to the number of words in the text times 300, which is very large, and this might result in poor performance. In contrast, logistic regression shows relatively good performance.

We chose the ensemble model as our best model because it outperformed all of the other models. So, in the next subsection, we'll discuss the performance of the ensemble model. 

\subsection{Discussion}
\begin{figure}[H]
    \subfloat[Accuracy of training and validation set over epoch]{
        \begin{minipage}[c][]{0.5\textwidth}
            \centering
             \includegraphics[width=1\textwidth]{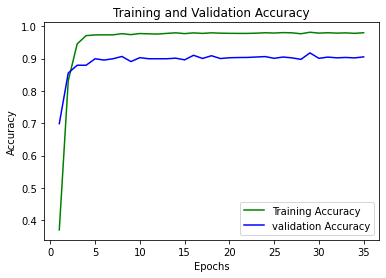}
        \end{minipage}
    }
    \hfill
    \subfloat[Loss on training and validation set over epoch]{
        \begin{minipage}[c][]{0.5\textwidth}
            \centering
            \includegraphics[width=1\textwidth]{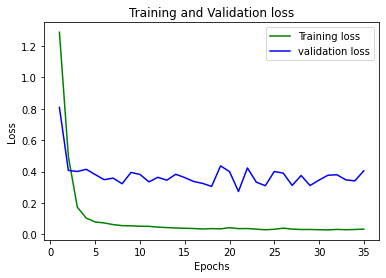}
        \end{minipage}
    }
    \caption{Accuracy and loss curve of the Ensemble model }
    \label{fig:accuracyloss}
\end{figure}

Figure \ref{fig:accuracyloss}(a) shows the training and testing accuracy of the ensemble model over the epochs  and figure \ref{fig:accuracyloss}(b) shows the training and testing loss of the ensemble model over the epochs.  We can see that the gap between the training and validation accuracy curve is not too high indicating that there is very little overfitting but not significant.

\begin{figure}
	\centering
		\includegraphics[scale=.70]{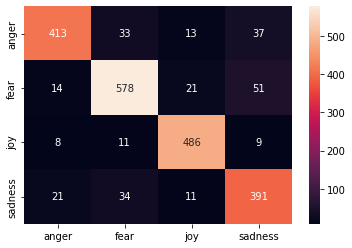}
	\caption{Confusion matrix of the ensemble model}
	\label{FIG:report2}
\end{figure}

Table \ref{tab:report1} and figure \ref{FIG:report2} show the classification report and the confusion matrix of the ensemble model respectively. The classification report contains a precision, recall, and f1-score for each of the classes. F-1 score is a metric for determining how accurate a classifier is. This score is calculated as a weighted average of precision and recall. According to the classification report, class sadness has the lowest f1-score when compared to other classes, and the confusion matrix explains the reason behind this. We can see that 51 samples from class fear and 37 samples from class anger are classified as sadness which produces a lower precision value for class sadness and contributes  to the lower f1-score. However, no class has an f1-score less than 80\% which indicates a satisfactory result. Besides, by observing the number of support(data sample) of each class from the classification report, we can say  that the more data a class has, the higher the f1-score of that class is. So by increasing the data, we can increase the f1-score of each class. Moreover, accuracy, macro average accuracy and weighted average accuracy are also satisfactory as expected.


\begin{table*}[]
    \centering
\caption{Classification report of the ensemble model}
\label{tab:report1}
\begin{tabular}{|c|c|c|c|c|}
\hline
                      & \textbf{Precision} & \textbf{Recall} & \textbf{F1-score} & \textbf{Support} \\ \hline
\textbf{Anger}        & 0.9057             & 0.8327          & 0.8676            & 496              \\ \hline
\textbf{Joy}          & 0.8811             & 0.8705          & 0.8758            & 664              \\ \hline
\textbf{Fear}         & 0.9153             & 0.9455          & 0.9301            & 514              \\ \hline
\textbf{Sadness}      & 0.8012             & 0.8556          & 0.8275            & 457              \\ \hline
\textbf{Accuracy}     &                    &                 & 0.8766            & 2131             \\ \hline
\textbf{Macro avg}    & 0.8758             & 0.8761          & 0.8753            & 2131             \\ \hline
\textbf{Weighted avg} & 0.8779             & 0.8766          & 0.8766            & 2131             \\ \hline
\end{tabular}
\end{table*}

    \begin{figure}[H]
      \centering
        \includegraphics[width=10cm]{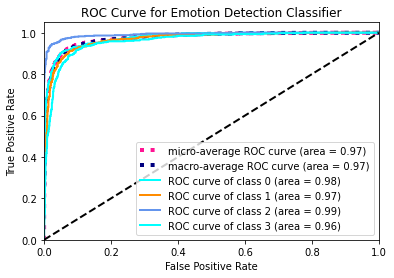}
      \caption{ROC curve of the ensemble model}
      \label{fig:roc_curve}
  \end{figure}
   The ROC curve for the ensemble model is shown in the Figure \ref{fig:roc_curve}. From this figure we can also see the ROC curve of each of the classes.
   In this ROC curve, class 0, class 1, class 2, and class 3 indicate classes anger, fear, joy, and sadness, respectively. It is observed that class 3(sadness) has the lowest area under the curve, which is 0.96. This is because, according to the classification report,  we have already observed that number of false positive rate is higher in class 3 (sadness) compare to other classes. However, the overall micro-average and macro average ROC curve area is around 0.97 which is expected. This means that the ensemble model performed well and fulfilled our expectation.

\section{Conclusion and Future Scope}
Social Media's are becoming our daily life activity. We spend a significant amount of time  on social media every day. Therefore, it's contents are important to understand the behavior of it's users. In this project we aim to classify the texts as four classes of emotion : Fear, Anger, Sadness and Joy. We have used several machine learning models as well as ensemble of two deep learning models to get the best performance. Our ensemble model has a high level of accuracy, and it may be used to offer informative content on a user's homepage depending on his or her previous behaviors.

In future, we want to reduce the gap between training and validation accuracy as well as to observe how our model performs as we enlarge the dataset. We do not include any transfer learning and graph based model in this project. We want to train our model with  large data set and want to deploy the learning in some transfer learning methods to see the performances. Graph based models are also an alternative option that we may work in near future.

\bibliographystyle{IEEEtran}
\bibliography{bibliography}

\end{document}